\documentclass[conference]{IEEEtran}

 \usepackage{pgfplots}
\usepackage{mathtools} 
\usepackage{multirow}
\usepackage[english]{babel}
\usepackage{float}
\usepackage{breqn}
\usepackage{footnote}
\usepackage{subfig}
\usepackage{balance}
\usepackage{blindtext, graphicx}
\usepackage{booktabs}
\usepackage{soul,color}
\usepackage{multirow}
\usepackage{algorithm,algorithmic}
\usepackage{bm}
\usepackage{ifpdf}
\usepackage{amsmath}
\usepackage{array}
\usepackage{multirow}

\ifCLASSINFOpdf

\else

\fi

\newcommand{\cmmnt}[1]{}

\begin{document}

\title{\huge{Processing-In-Memory Acceleration of Convolutional Neural Networks for Energy-Efficiency, and Power-Intermittency Resilience}}\vspace{-1.2em}

\author{Arman Roohi\textsuperscript{$\ast$}, Shaahin Angizi\textsuperscript{$\ast$}, Deliang Fan, and Ronald~F~DeMara\\
Department of Electrical and Computer Engineering, University of Central Florida, Orlando, 32816 USA\\
\small
\textsuperscript{$\ast$}The first two authors contributed equally
\normalsize
\vspace{-0.8em}
}

\maketitle

\begin{abstract} 
Herein, a bit-wise Convolutional Neural Network (CNN) in-memory accelerator is implemented using Spin-Orbit Torque Magnetic Random Access Memory (SOT-MRAM) computational sub-arrays. It utilizes a novel AND-Accumulation method capable of significantly-reduced energy consumption within convolutional layers and performs various low bit-width CNN inference operations entirely within MRAM. Power-intermittence resiliency is also enhanced by retaining the partial state information needed to maintain computational forward-progress, which is advantageous for battery-less IoT nodes. Simulation results indicate $\sim$5.4$\times$ higher energy-efficiency and 9$\times$ speedup over ReRAM-based acceleration, or roughly $\sim$9.7$\times$ higher energy-efficiency and 13.5$\times$ speedup over recent CMOS-only approaches, while maintaining inference accuracy comparable to baseline designs.

\end{abstract}

\IEEEpeerreviewmaketitle

\section{Introduction}

Due to their impressive performance on image recognition tasks, deep Convolutional Neural Network (CNNs) offer significant potential advantages for use on large-scale data-sets. However, the processing demands of high-depth CNNs spanning hundreds of layers face serious challenges for their tractability in terms of memory and computational resources. This so-called ''CNN power and memory wall'' has been motivating the development of alternative approaches to improve CNN efficiency at both software and hardware levels \cite{andri2017yodann}. 

In algorithm-based approaches, use of shallower CNN models, quantizing parameters \cite{zhou2016dorefa}, and network binarization \cite{rastegari2016xnor} have been explored extensively. Recently, utilizing weights with low bit-width and activations reduces both model size and computing complexity. For instance, performing bit-wise convolution between the inputs and low bit-width weights has been demonstrated in \cite{zhou2016dorefa} by converting conventional Multiplication-And-Accumulate (MAC) operations into their corresponding AND-bitcount operations. However, such conversion cannot necessarily guarantee high efficiency operation in a hardware implementation that may engage various aspects of instruction encoding and operand access. In an extreme quantization, Binary Convolutional Neural Network (BCNN) has achieved acceptable accuracy on both small \cite{courbariaux2016binarized} and large datasets \cite{rastegari2016xnor} by relaxing the demands for some high precision calculations. Instead, it binarizes weight/input while processing the forward path, providing a promising solution to mitigate aforementioned bottlenecks in storage and computational components \cite{angizi2017imc}.   

From the hardware point of view, the underlying operations should be realized using efficient mechanisms. However, within conventional isolated computing units and memory elements interconnected via buses, there are serious challenges, such as limited memory bandwidth channels, long memory access latency, significant congestion at I/O chokepoints, and high leakage power consumption \cite{chi2016prime,imani2017mpim}. In-memory processing paradigms built on top of non-volatile devices, such as Resistive Random Access Memory (ReRAM) \cite{chi2016prime,tang2017binary}, Spin-Transfer Torque Magnetic RAM (STT-MRAM) \cite{fong2016spin} and recent Spin Orbit Torque MRAM (SOT-MRAM) \cite{he2017exploring}, introduced to address the aforementioned concerns. Due to their interesting features such as non-volatility, near-zero standby power, high integration density, compatibility with CMOS fabrication process, and radiation-hardness, these NV-based systems offer some promising attributes for in-memory processing implementations.

\begin{figure}[t]
\begin{center}
\begin{tabular}{c}
\includegraphics [height=4cm]{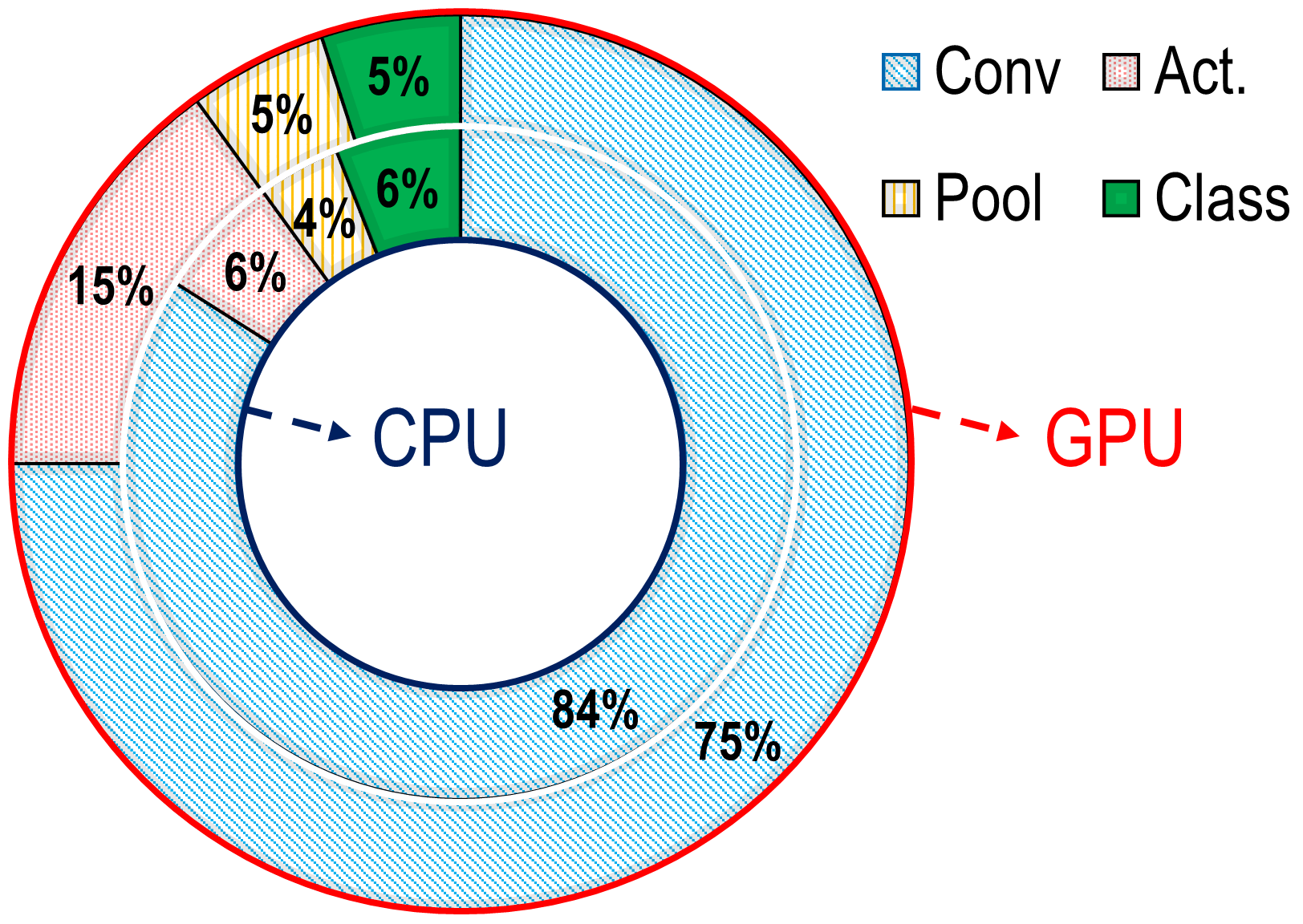}\\
 \end{tabular}
\vspace*{-1em}
\caption{Proportional relationship for execution time of a CNN on both CPU and GPU \cite{cavigelli2015accelerating}.}\vspace{-2em}
\label{GPU} 
\vspace*{-1em}
\end{center}
\end{figure}

CNNs realize machine learning classifiers that are capable of taking an image as an input and then computing the probability that the image belongs to each designated output class. Typically, a CNN consists of several convolutional layers including convolution, non-linearity, normalization, and pooling steps, followed by a flatten layer connected to fully-connected layers.
For feature extraction, each convolutional layer receives a set of features organized into multi-channels referred to as \textbf{feature maps}. It applies \textbf{feature detectors}(filters) by performing high-dimensional convolutions. To increase non-linearity of the pooled feature map, a non-linear activation function, i.e. rectified linear unit (ReLU), will be applied to the results. The convolutional layer occupies the largest portion of running time and consumes significant computational resources in both GPU and CPU implementations, as depicted in Figure \ref{GPU}. This motivates us to propose an optimized bit-wise CNN in-memory accelerator based on SOT-MRAM computational sub-arrays. In particular, the bit-wise CNN based on AND-bitcount operations presented in \cite{zhou2016dorefa} can be further accelerated by modifying the algorithm rather than a direct module-by-module mapping such as IMCE \cite{angizi2018imce}.

\cmmnt{
The main contributions of this work are summarized as follows: (1) The proposed accelerator is designed to be partially power-failure resilient and uses a novel hardware-conforming AND-Accumulation method to further accelerate convolutional layers in CNN; (2) a computational sub-array architecture based on SOT-MRAM is exploited to develop a bit-wise CNN in-memory accelerator;  (3) after presenting a  detailed hardware mapping of low bit-width CNNs for the proposed design, we perform extensive simulations with the proposed CNN accelerator, such as inference accuracy, memory storage, energy consumption, and performance to assess its performance.
}
The remainder of this paper is organized as follows. In Section II, the proposed accelerator is designed to be partially power-failure resilient and uses a novel hardware-conforming AND-Accumulation method to further accelerate convolutional layers in CNN. The non-volatile SOT-MRAM based structures provide
power failure resiliency feature. Moreover, they are leveraged to develop a bit-wise CNN in-memory accelerator. Extensive simulation results and detailed analysis are summarized in Section III including inference accuracy, energy consumption, and memory storage. Finally, Section IV concludes this paper by highlighting the features and advantages of the proposed in-memory accelerator.

\cmmnt{\section{In-Memory Computational Sub-array}}
\section{Intermittent Resilient CNN Accelerator}
\begin{figure}[t]
\begin{center}
\begin{tabular}{c}
\includegraphics [width=0.95\linewidth, height=6.3cm]{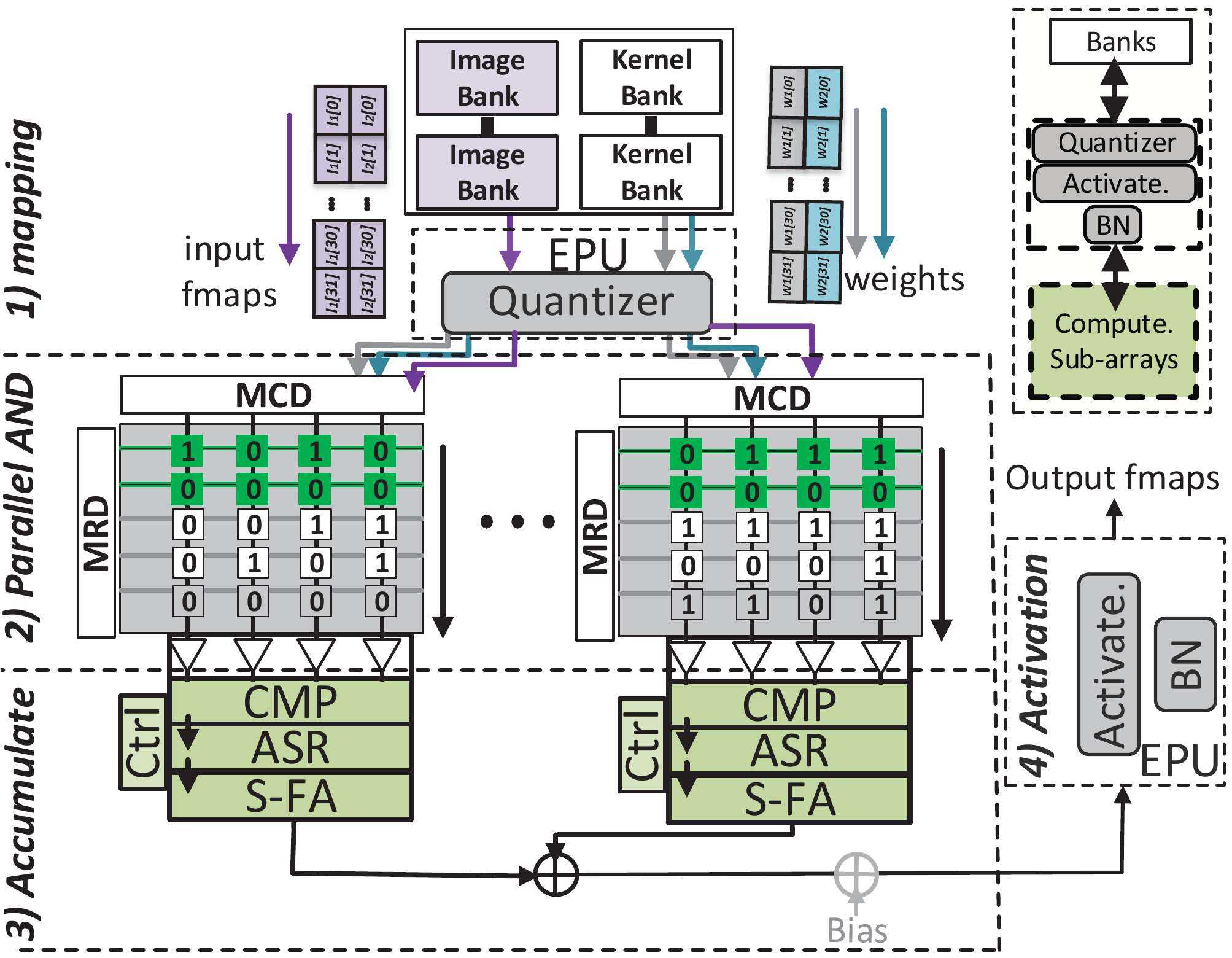}\\
 \end{tabular}\vspace{-0.8em}
\caption{General overview of the proposed intermittent resilient CNN accelerator.}\vspace{-2.5em}
\label{arc}
\end{center}
\end{figure}

\cmmnt{\section{Accelerator Architecture}}
As mentioned in the previous section, most of the CNN run-time is taken by performing MACs. Therefore, reducing computation complexity of MAC operations are vital for resource-constrained systems such as IoT devices. In order to accelerate MAC operations in convoltional layers, three main processes including \textit{AND} operation, \textit{bitcount}, and \textit{bitshift} are leveraged, which realize  a bit-wise convolution.\cmmnt{in DoReFa-Net \cite{zhou2016dorefa}.} 
\cmmnt{The key idea behind employing bit-wise convolution in DoReFa-Net \cite{zhou2016dorefa} is to exploit \textit{logic AND}, \textit{bitcount}, and \textit{bitshift} as rapid and parallelizable operations to accelerate the MACs in convolutional layers. As these operations are polynomial in the product of bit-width multiplicands, reducing the bit-width brings remarkable improvements in computation complexity dominating the run-time of CNN, specially in resource-constrained environments.} 
A crude in-memory implementation of such bit-wise operations can be found in \cite{angizi2018imce}, where  \textit{bitcount} and \textit{bitshift} are directly implemented using serial counter and shifter units. We believe such module-by-module mapping not only degrades the bit-wise convolution performance in hardware, but also imposes a large in-memory data-transfer due to its intrinsic serial operations. 
Hence, we propose a hardware-optimized method inspired by DoReFa-Net \cite{zhou2016dorefa} to mitigate these drawbacks, in addition to address the power-failure issue:\cmmnt{which  to compute the dot-product and consequently the convolution of $k$-bit fixed point integers, i.e. input ($I$) and weight ($W$):} \vspace{-1em}
\small 
\begin{multline}
I\ast W=\sum _{m=0}^{M-1}\sum _{n=0}^{N-1}2^{m+n}CMP(AND(C_n(W),C_m(I)))\\
CMP(X)=\sum _{i=1}^{n}x_{i}, \text{ , where }  X= x_{n}x_{n-1}...x_{2}x_{1}
\vspace{1em}
\end{multline}
\normalsize
where $I$ as input and $W$ weight. The convolution can be implemented by \textit{AND}, \textit{CMP} (rather than \textit{bitcount}), and parallel \textit{bitshift} operations. A general overview of our proposed CNN accelerator is shown in Figure \ref{arc}a. This architecture mainly consists of an Image Bank, a Kernel Bank, computational sub-arrays, and an Extra Processing Unit (EPU)  including three  ancillary  units, i.e. Quantizer, Activation Function-Active, and Batch Normalization-BN. Each computational sub-array is equipped with three components: CMP as a Compressor unit, ASR as an Adaptive Shift Register, and NV-FA as a Non-Volatile Full Adder. 
As discussed earlier, the convolutional layer contributes the largest proportion of computation time and complexity to CNNs. Thus, we mainly focus on this layer. However, the proposed system architecture can be leveraged to implement other CNN layers including batch normalization and pooling layers. 
We assume both feature maps ($I$) and feature detectors ($W$) are initially loaded in two sub-banks of memory.
In our approach, inputs should be quantized before mapping to the accelerator, which is performed by EPU's Quantizer.
Because of page limit of this paper we cannot thoroughly discuss EPU's units in details. 
In the following, we elaborate two main processing phases of the accelerator. 


\cmmnt{\subsection{Mapping Phase}}

\begin{figure}[b]
\vspace{-1em}
\begin{center}
\begin{tabular}{c}
\includegraphics [width=1\linewidth, height=5.75cm]{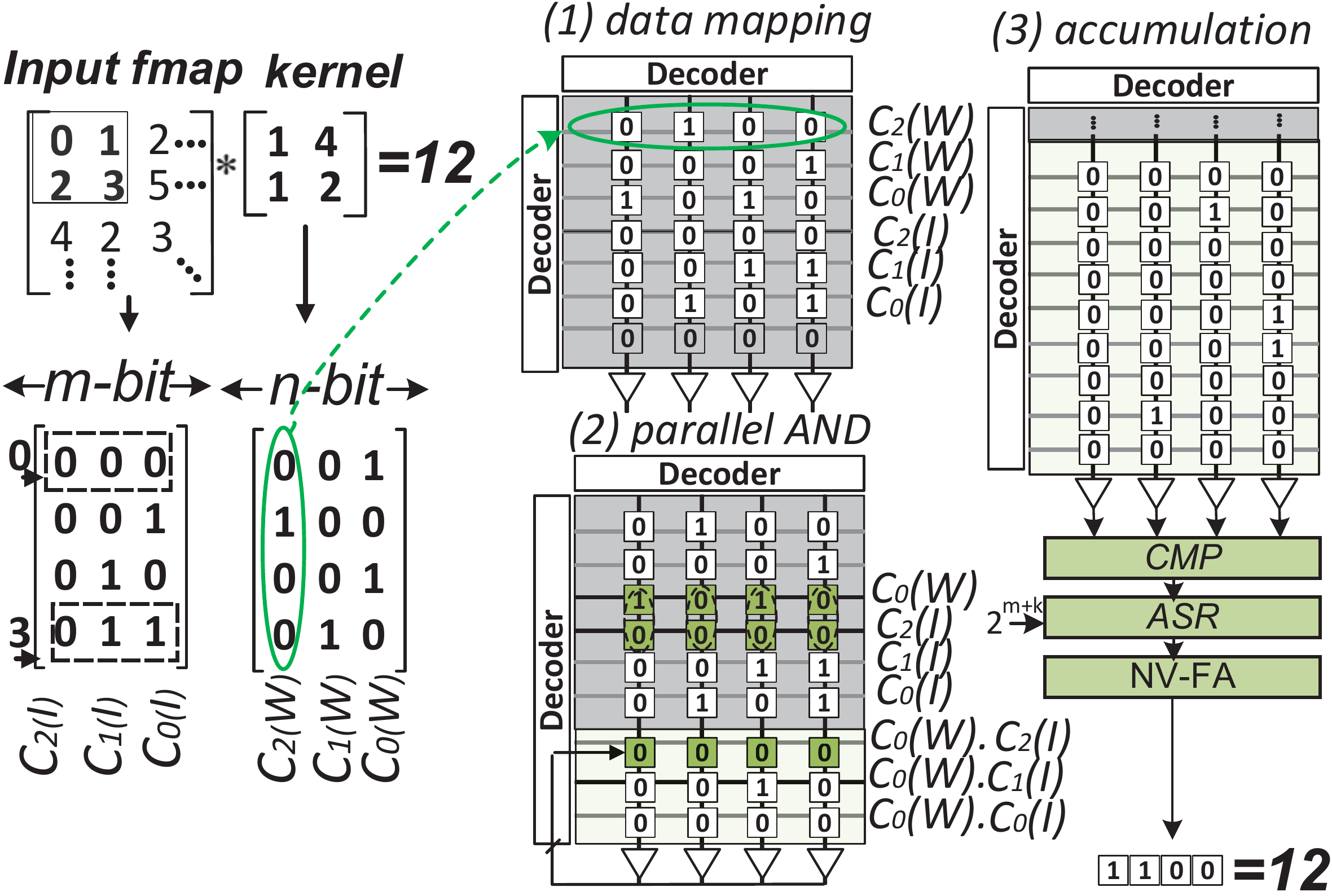}\\
 \end{tabular}\vspace{-1em}
\caption{Three-phase in-memory computation.}\vspace{-0.2em}
\label{map}
\end{center}
\end{figure}

\subsection{Parallel AND Phase}\vspace{-0.2em}

\begin{figure}[b]
	\begin{center}
		\begin{tabular}{c}
			\includegraphics [width=0.95\linewidth]{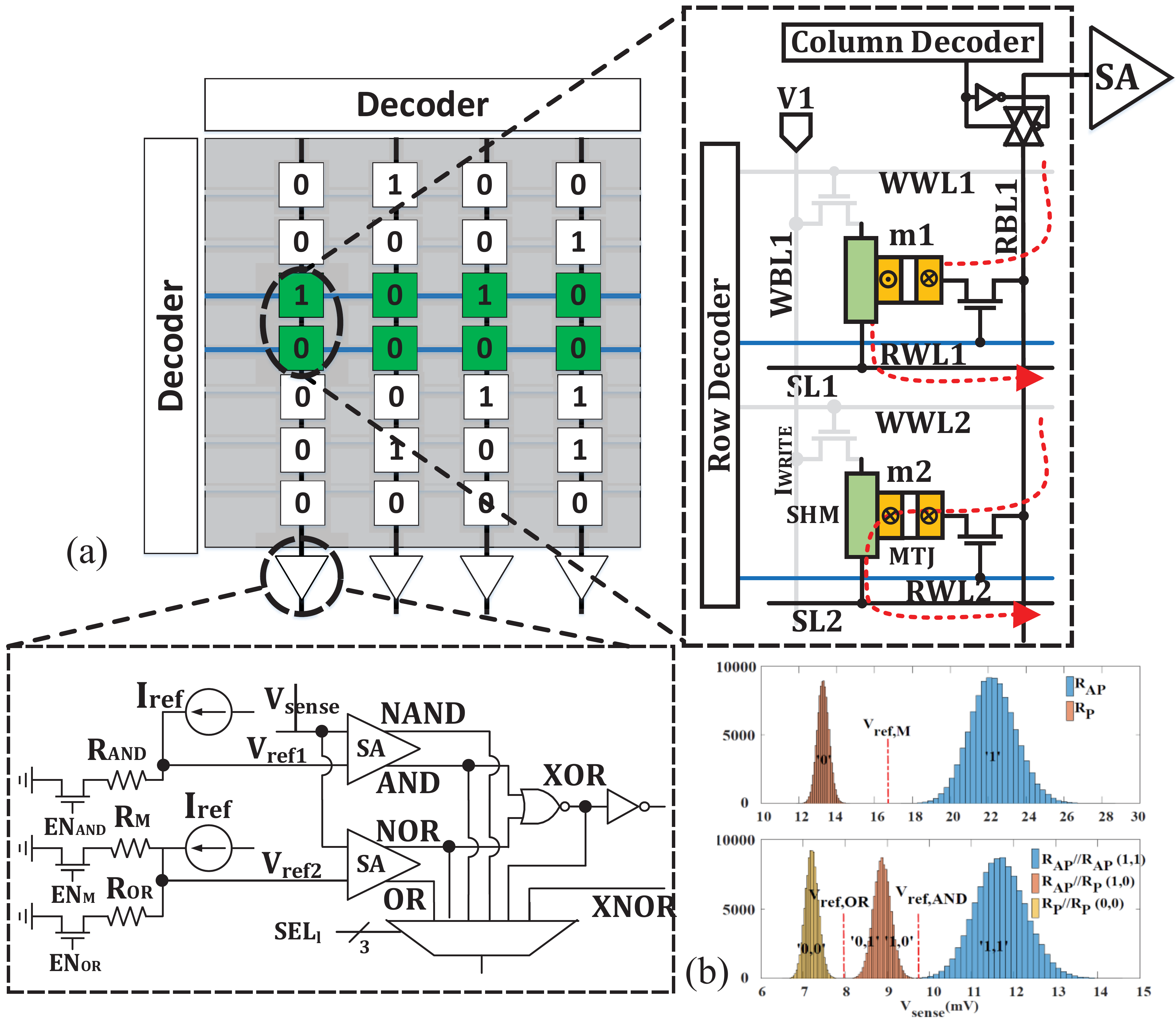}\\
			\hspace{0.8cm} 
		\end{tabular}
		\vspace{-1em}
		\vspace{-0.8em}
		\caption{(a) SOT-MRAM Computational sub-array, (b) Monte Carlo simulation result of V\textsubscript{sense}.}
		\vspace{-2em}
		\label{Array} 
	\end{center}
\end{figure}

Figure \ref{Array}a shows the in-memory processing sub-array architecture using SOT-MRAM  \cite{he2017exploring,angizi2018imce,Angizi2018CMP}. The array supports both memory read-write and simple Boolean logic operations such as AND/XOR.
The SOT-MRAM structure includes an Magnetic Tunnel Junction (MTJ) that its free layer is directly connected to a Spin Hall Metal (SHM). 
\cmmnt{The resistance of MTJ with parallel magnetization in both magnetic layers (data-`0') is lower than that of MTJ with anti-parallel magnetization (data-`1').}
There  are  two  stable magnetization states, parallel (low resistance), and anti-parallel (high resistance), which denote “0”  and  “1”  in  binary  information,  respectively. 
Each SOT-MRAM cell requires five signals, which are common among all MRAM cells to perform memory operations. There are Write Word Line (WWL), Write Bit Line (WBL), Read Word Line (RWL), Read Bit Line (RBL), and a Source Line (SL). (For more details, refer to \cite{Roohi2018TC}\cmmnt{, in which an extensive analysis is studied.})

The SOT-MRAM based computational sub-array can be readily utilized such that the massive AND operations required for convolutions can be handled. Consider $I$ and $W$ as input and kernel of $m$- and $n$-bit (for simplicity, 3-bit, as Figure \ref{map}), $I$ is covered by kernel $W$. The bits of each $I_i$/ $W_i$ element are indexed from least significant bit to moat significant bit with $M=[0 , m-1]$/$N=[0 , n-1]$. Then, a second sequence noted by $C_m(I)$ can be considered for $I$ including the combination of $m^{th}$ bit of $I_i$ elements. For example, $C_2(I)$ represents the LSBs of all $I_i$ elements, ``0000". The second sequence for $W$ can be considered like $C_n(W)$. Now, by considering the set of all $m^{th}$ value sequences, the $I$ can be expressed as $I=\sum_{m=0}^{M-1}2^mC_m(I)$. Additionally, $W$ can be expressed as $W=\sum_{n=0}^{N-1}2^nC_n(W)$.

To efficiently load the Quantizer unit's output to computational sub-arrays, $I$ and $W$ should be tailored. As illustrated in the data organization and mapping step of Figure \ref{map}, $C_2(W)$-$C_0(W)$ are consequently mapped to the assigned sub-array. Accordingly, $C_2(I)-C_0(I)$ are mapped to the following memory rows similarly. Now, the accelerator can perform the parallel bit-wise AND operation depicted in Figure 4 within its computational sub-array.

\subsection{Accumulation Phase}
The accumulation phase consists of three main components: (1) NV 4:2 compressor, (2) adaptive shift register, and (3) NV full adder. 
\subsubsection{4:2 Compressor (CMP)}

\begin{figure}[t]
\begin{center}
\begin{tabular}{c}
\includegraphics [width=0.98\linewidth, height=2.3cm]{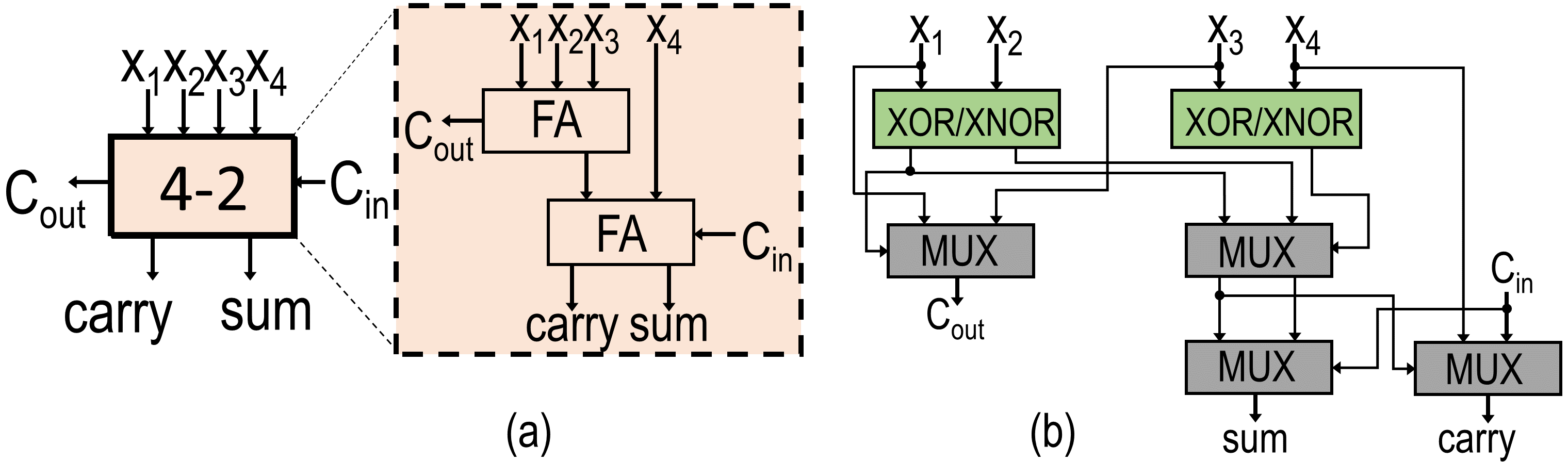}\\
 \end{tabular}\vspace{-0.8em}
\caption{Implementation of (a) 4:2 compressor, and (b) proposed architecture with the MUX and XOR-XNOR modules.}
\label{comp}
\end{center}
\vspace{-2em}
\end{figure}

Compressors, especially 4:2 and 5:2, are widely used to reduce the delay of the summation of partial products in multiplier designs. Figure \ref{comp}a shows the schematic of a 4:2 compressor and its fundamental implementation using two serially connected full adders. The basic equation of the 4:2 compressor is x\textsubscript{1} + x\textsubscript{2} + x\textsubscript{3} + x\textsubscript{4} + C\textsubscript{in} = sum + 2 $\times$ (carry + Cout). The following equations express the outputs of the 4:2 compressor:\vspace{-2em}

\begin{multline}
sum = \rm x_{1} \oplus x_{2} \oplus x_{3} \oplus x_{4} \oplus c_{in}\\
carry = (\rm x_{1} \oplus x_{2} \oplus x_{3} \oplus x_{4} ).c_{in} + \overline{(x_{1} \oplus x_{2} \oplus x_{3} \oplus x_{4})}.x_{4}\\
Cout = (\rm x_{1} \oplus x_{2}). x_{3} + \overline{(x_{1} \oplus x_{2})}.x_{1}
\end{multline}

These equations can be reformed in a way that XOR/XNOR modules are only located in a first row and the other XOR/XNOR elements are replaced by MUXs, as shown in Figure \ref{comp}b. Considering the compressor implementation presented in \cite{momeni2015design} and the capability of our proposed computational sub-array design to function as an XOR/XNOR operation including MUX elements, the accelerator can be configured to implement an optimized 4:2 compressor. 
The results of parallel AND operations are written back to the sub-array and passed through the compressor, which can readily count the number of ``1''s within each resultant vector and pass it to the next unit. Figure \ref{comp}b depicts step (1) of the accumulation phase. In our design, we only need to update the memory contents once to implement XOR/XNOR logic, namely in-memory XOR computation. Due to the 4:2 compressor, the bitcount operation can be performed in one clock cycle instead several \footnote{The number of shift is determined by the memory array size, i.e. 8 bits.}  clock cycles of shifting operations, yielding considerable reductions in delay and energy. Due to the non-volatile XOR/XNOR implementation, our 4:2 compressor is power failure resilient. Moreover, it is power efficient, due to the optimum number of write operations equal to the sub-array length. In general, kernel length ($\rm n_k$) determines the number of compressor's input ($\rm n+1$).

\subsubsection{Adaptive Shift Register (ASR)}
\begin{figure}[b]
\begin{center}
\begin{tabular}{c}
\includegraphics [width=0.99\linewidth, height=5.5cm]{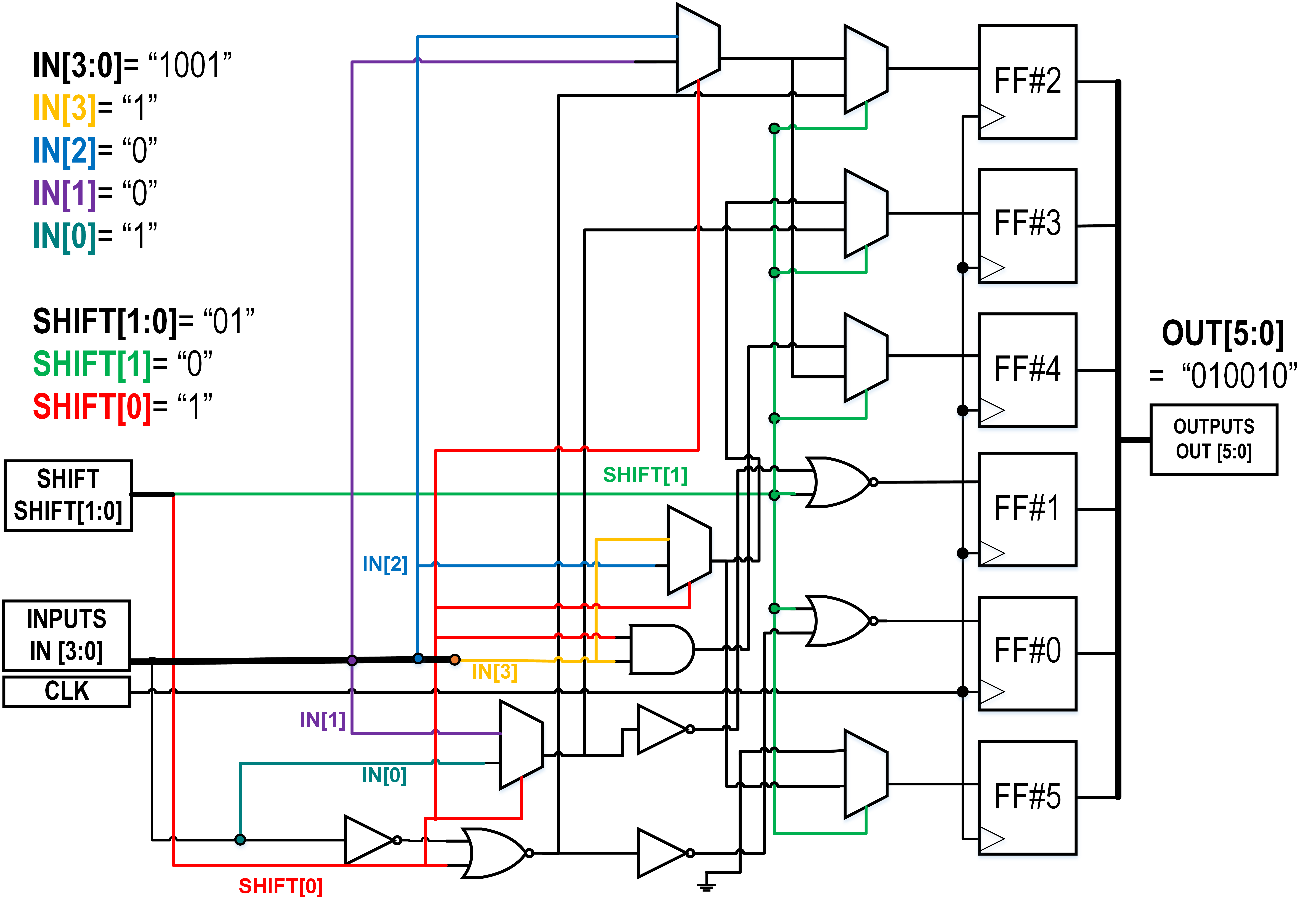}\\
 \end{tabular}
\caption{4-bit adaptive shift register with three shift modes.}
\label{shift}
\end{center}\vspace{-2.5em}
\end{figure}

Since the number of shift operations is different and is determined by the locations of the input and the weight in the sub-array, governed by the expression: $\rm m+n-2$, an adaptive shift register (ASR) is required. One method to implement an ASR is an addition tree approach. In general, this structure is composed of $\rm 2^{m+n}-1$ bit full adders (FAs), in which the first layer includes $\rm 2^{n-1}$ FAs, the second layer has $\rm 2^{n-2}$ FAs, and finally the last layer consists of one FA. Another approach to designing an ASR, developed herein, is to implement logic expressions using multiplexers (MUXs) and then connect them to the flip-flops (FFs) in an appropriate way. Figure \ref{shift} depicts an ASR design for 4-bit input data, which is able to operate with three different numbers of shifts, 00=0, 01=1, and 10=2. It includes seven MUXs, three inverters, four NOR/AND gates, and six FFs. For instance, assume that IN[3:0] = ``1001'' and SHIFT[1:0] is 1 (01), which means SHIFT[0]=1 (red line) and SHIFT[1]=0 (green line). Because of the MUX-based selection structure, ``0'' is stored in FF\#5 and FF\#0. Then the applied input is written into FF\#1 to FF\#4 appropriately/successively, which produces ``010010'' as an output. 
The number of FFs is determined by the summation of the number of inputs and the maximum number of possible shift operations. In our implementation, a 4-bit ASR is developed, which requires six FFs to perform three possible shift operations.

\subsubsection{Non-Volatile Full Adder (NV-FA)}
\begin{figure*}[h]
\begin{center}
\begin{tabular}{c}
\includegraphics [width=.85\linewidth, height=4.25cm]{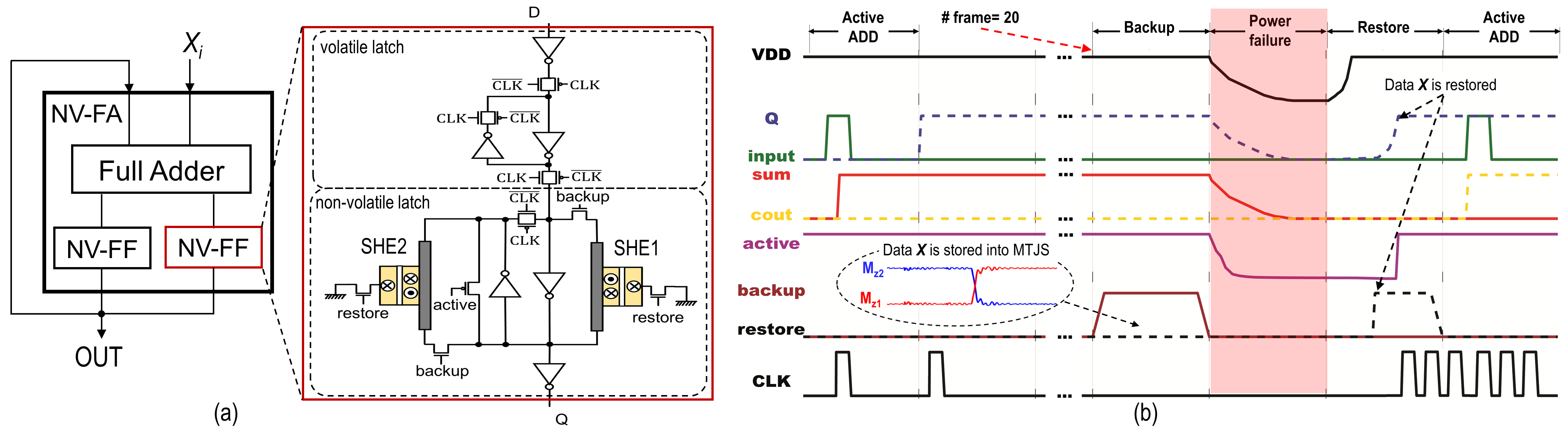}\\
 \end{tabular}
\caption{(a) Circuit level design of proposed NV-FA, and
(b) timing diagram for NV-FA operation.}\vspace{-2.2em}
\label{NVFA}
\end{center}
\end{figure*}

Due to the usage of NV elements in the AND-Accumulation process, the structure has become partially power-failure resilient, meaning it can restore the system's operations to the last good/suitable state under most conditions. While it might fail to restore the last configuration if power loss occurs during the addition (shift) operations, the delay for this step is equal to the delays of m+n FAs, $\rm \approx m+n\times 58$ ps, which is negligible to the total delay of calculating one fmap. Finally, the output of this level should be added with the results of the previous inputs (Is $\times$ Ws). To make our design more resilient in presence of power failure, we developed a NV-FA (NV-FA), as shown in Figure \ref{NVFA}a. The NV-FA includes two NV flip-flops (NV-FFs) in addition to the regular FA. The NV-FF consists of a volatile CMOS FF and a NV element.

To remove the additional overhead and issues caused by the common checkpointing approaches, the summation results will be written into the NV-elements after computing steps for a fixed number of frames, i.e. 20 frames. Otherwise, the states and results of each step store in a volatile FF and sum up with the upcoming results. We can modify the period of writing operation based on the power failure rate. In this case, our checkpointing approach is superior to common energy harvesting systems, which are usually utilized in intermittent computing architectures, in terms of area and complexity. Because herein, common checking-based approaches may suffer from inconsistencies, both internal and external, after each power loss. Moreover, peripheral circuits such as voltage detection systems and capacitor arrays are needed, which is a crucial challenge for area-constrained IoTs. Figure \ref{NVFA}b depicts the functionality of NV-FA in presence of power failure.

\cmmnt{
\subsection{EPU}

\indent\textit{1) Quantizer:}
This component quantizes a real number input $r_i$ $\in$ [0, 1] to a $k$-bit number output $r_o$ $\in$ [0, 1] using quantization function \cite{zhou2016dorefa}: 
\begin{equation}
r_o=\frac{1}{2^k-1}round((2^k-1)r_i)
\end{equation}

\indent\textit{2) BN:}
Batch Normalization layer \cite{zhao2017accelerating} alleviates the information loss during quantization by normalizing the input batch to have zero mean and unit variance. The transformation can be written as:
\begin{equation}
I_o(R)=\frac{I_i(R)-\mu}{\sqrt{\sigma^{2}+\varepsilon}}\gamma +\beta  
\end{equation}
where $I_o(R)$ and $I_i(R)$ denote the corresponding output and input pixels, respectively. $\sigma$ and $\mu$ represent statistics achieved during training mode, $\gamma$ and $\beta$ are trained parameters, and $\varepsilon$ is included to avert round-off problem. During inference mode, all the parameters in equation (4) are stored in SOT-MRAM sub-arrays, so the EPU efficiently fetches each pixel of input fmap and writes back the corresponding normalized pixel.  

\indent\textit{3) Active:}
After performing the accumulation operation, the generated fmap needs to be activated by employing an activation function (see Figure \ref{arc}). The proper selection of the activation function has a profound impact on network prediction accuracy in low-bitwidth CNNs. In our accelerator, this unit can be adjusted to perform two distinct functions (i.e. $ \frac{tanh(x)+1}{2}$ and $sign(x)$) to provide maximum accuracy.
}

\section{Experimental Results} 

\subsection{Accuracy}
\underline{Bit-width:} We consider 4 different bit-width of W:I (1:1, 1:4, 1:8, 2:20 to explore the accuracy of our accelerator with an 8-bit gradient. In addition, we consider a 32-bit full-precision case as the base-line with 32-bit gradient. \underline{Data-set:} Among various data-sets, we select SVHN \cite{netzer2011reading} with 73257 training digits, 26032 testing digits, and 531131 additional digits for extra training data. The images are pre-processed to 40$\times$40 from the original 32$\times$32 cropped version and fed to the model. \underline{CNN Layers:} We developed a bitwise CNN with 6 convolutional, 2 average pooling and 2 FC layers, which are equivalently implemented by convolutional layers. Such model costs about 80 FLOPs for each 40$\times$40 image. To avert further prediction accuracy degradation, we don't quantize the first and last layers \cite{zhou2016dorefa,rastegari2016xnor,angizi2017imc}. \underline{Training:} We basically modified the open-source DoReFa-Net \cite{zhou2016dorefa} algorithm by integrating new bit-wise convolution function applying the AND-Accumulation method. To increase the accuracy and avoid over-fitting, we adopted batch normalization, parameter tuning and dropout methods. The CNN design is implemented on TensorFlow \cite{abadi2016tensorflow} running 100 epochs and we extract the test error of each epoch. \underline{Results:} Table \ref{acc} shows the computation complexity and test error of the under-test model. We used $W\times I$ and  $W\times I + W\times G$ to achieve the computation complexity of inference and training, respectively. 
Our results replicate the conclusion drawn by \cite{zhou2016dorefa,angizi2018imce} whereby kernels weights and inputs are progressively more vulnerable to bit-width reductions. 

\begin{table}[h]
\begin{center}
\caption{Test error of the CNN model on SVHN.} 
\scalebox{0.95}{\begin{tabular}{ccccc}
\hline
\multicolumn{2}{c}{Bit-width} & \multicolumn{2}{c}{Computation Complexity} & \multicolumn{1}{c}{Error (\%)} \\ \hline
W               & I               & Inference      & Training      & CNN Model \\      
32              & 32              & -   & -             & 2.4   \\      
1               & 1               & 1              & 9             & 3.1      \\     
1               & 4               & 4              & 12            & 2.3  \\       
1               & 8               & 8              & 16            & 2.1  \\       
2               & 2               & 4              & 20            & 1.8    \\ \hline     
\end{tabular}}
 \vspace{-1em}
\label{acc}
\end{center}
\end{table}

\vspace{-0.5em}
\subsection{Storage}
\begin{figure}[h]
\begin{center}
\begin{tabular}{cc}
\includegraphics [width=0.45\linewidth]{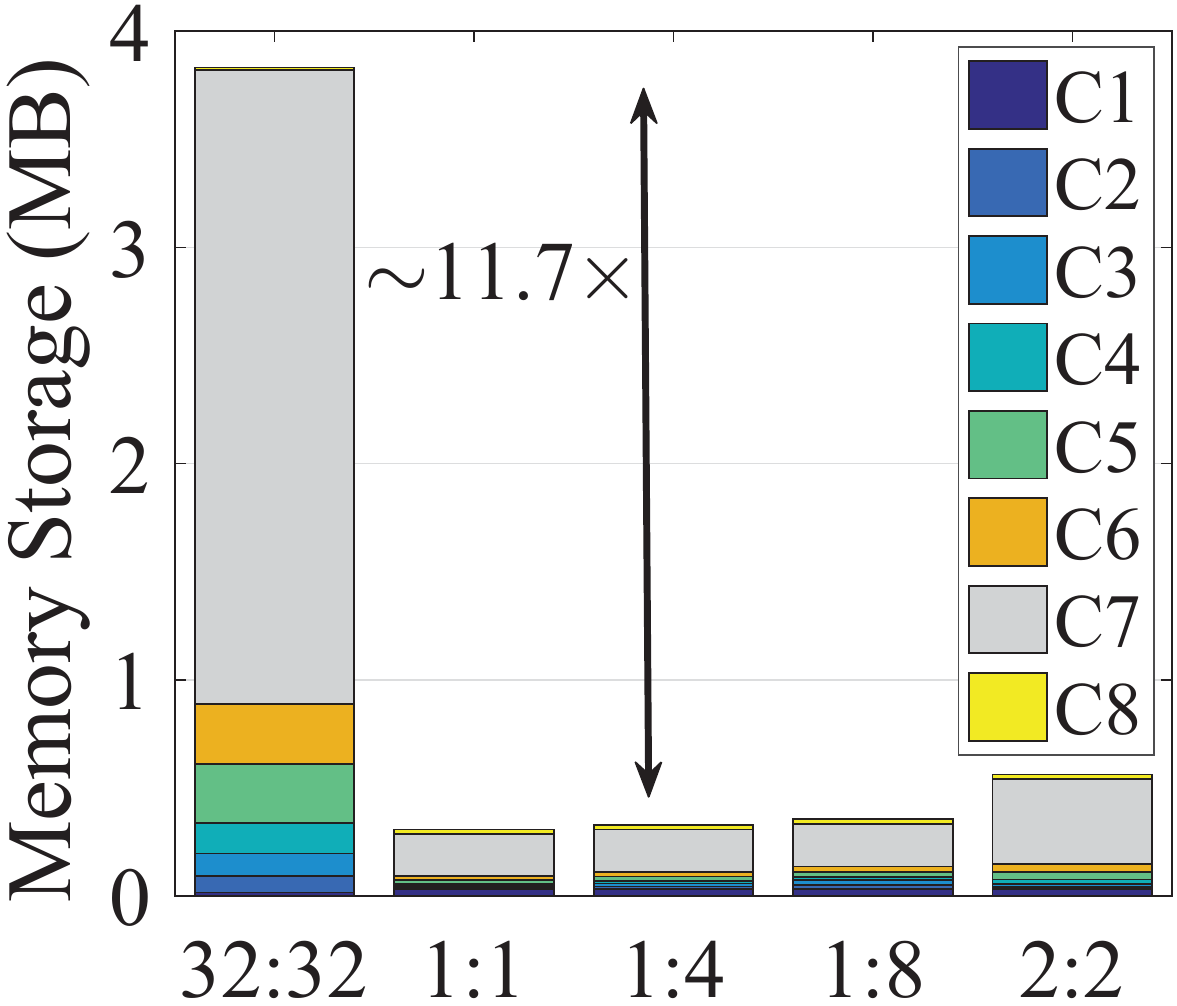}&\includegraphics [width=0.45\linewidth]{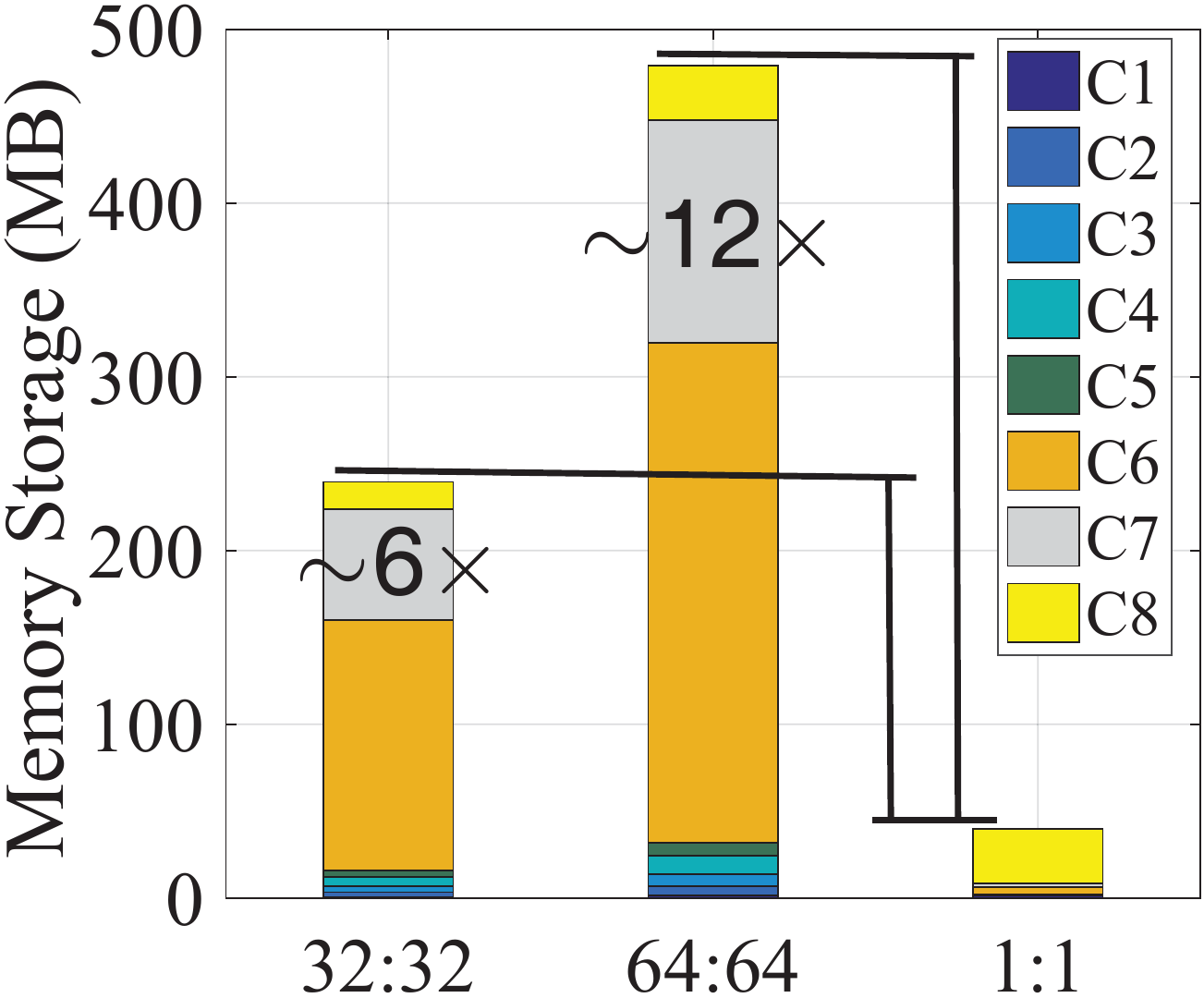}  \\ 
\small (a) &\small (b)
 \end{tabular}
\caption{Usage of memory storage by (a) the CNN model for SVHN date-set, and (b) AlexNet for ImageNet data-set.}
\label{mem_sto}\vspace{-1.8em}
\end{center}
\end{figure}

Five bit-width of W:I (32:32, 1:1, 1:4, 1:8, and 2:2) are selected to evaluate memory storage requirements. The breakdown of memory storage is shown in Figure \ref{mem_sto}a. We observe that as the CNN model's bit-width decreases, less memory storage is required. For instance, the 1:4 configuration, with higher inference accuracy compared to 32:32, shows $\sim$11.7$\times$ memory reduction. To investigate the memory usage in large data-sets, we implement three different bit-width of W:I (64:64, 32:32, and 1:1) using AlexNet model on ImageNet data-set on the proposed accelerator in \ref{mem_sto}b. We observe that 1:1 bit-width configuration demands $\sim$40MB memory which is $\sim$6$\times$ and $\sim$12$\times$ smaller in comparison with the single and double precision CNNs, respectively. 

\subsection{Energy Consumption}

In this subsection, we estimate the energy-efficiency of the CNN model implemented by the proposed accelerator and state-of-the-art inference acceleration solutions, i.e. ReRAM, SOT-MRAM, and CMOS-only ASIC. To evaluate the performance of the proposed design, the circuit level simulation is implemented in Cadence Spectre using NCSU 45nm CMOS PDK \cite{NCSU_PDK} in conjunction with a SOT-MRAM resistive model. The NEGF approach is utilized to extract the MTJ resistance (R\textsubscript{MTJ}) \cite{panagopoulos2012framework}, whereas the heavy metal resistance (R\textsubscript{SHM}) is determined based on the resistivity and device dimension. Accordingly, we extensively modified the system-level memory evaluation tool NVSim \cite{dong2014nvsim} to co-simulate with an in-house developed C++ code simulator based on circuit-level results. We configure the memory sub-array organization with 256 rows and 512 columns per mat organized in a H-tree routing manner, 2$\times$2 mats per bank, 8$\times$8 banks per group; in total 16 groups and 512Mb total capacity.

For comparison, a ReRAM-based in-memory accelerator based on \cite{chi2016prime} was developed with 64 fully-functional sub-arrays. For each mat, there are 256$\times$256 ReRAM cells and eight 8-bit reconfigurable SAs. For evaluation, NVSim simulator \cite{dong2014nvsim} was modified to estimate the system energy and performance. We adopted the default NVSim's ReRAM cell file (.cell) for the assessment. Besides, we developed an IMCE-like \cite{angizi2018imce} design with the same sub-array configuration as our design.
To compare the result with ASIC accelerators, we developed a YodaNN-like \cite{andri2016yodann} design with 8$\times$8 tiles for 33MB eDRAM. Accordingly, we  synthesized the design with Design Compiler \cite{DC} under a 45 nm process node. The eDRAM and SRAM performance were estimated using CACTI \cite{chen2012cacti}. 
In order to have a fair comparison, the area-normalized results (performance/energy per area) will be reported henceforth.

\begin{figure}[b]
	\centering
	\includegraphics[width=0.47\textwidth]{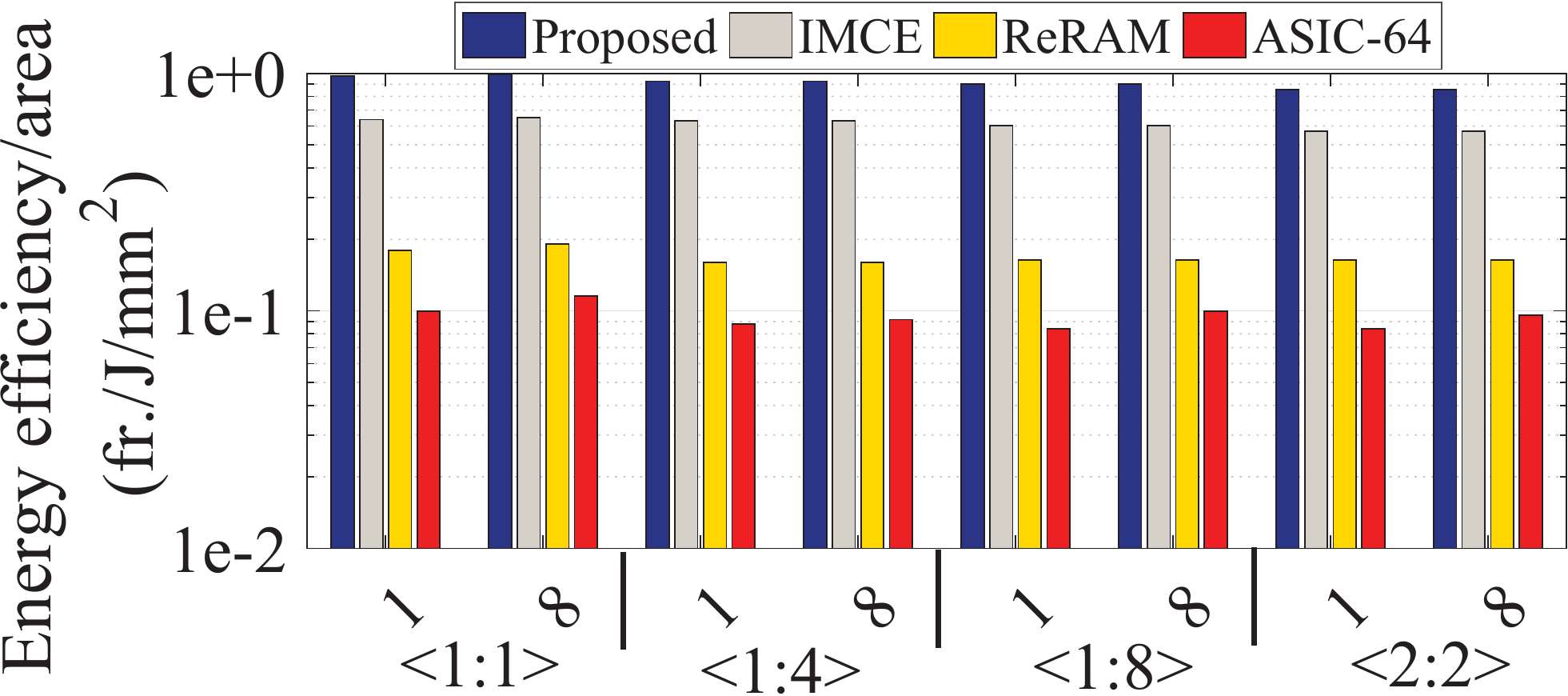}
	\caption{Energy-efficiency of different accelerators (Y-axis=Log scale).}
	\label{energy}	
\end{figure}

Figure \ref{energy} demonstrates the proposed accelerator energy-efficiency results with batch sizes of 1 and 8 in different configuration spaces of weight and input. We observe that the proposed accelerator offers the highest energy-efficiency normalized to area compared to others owing to its fast, energy-efficient, and parallel operations. Our design shows $\sim$2.1$\times$ better energy-efficiency compared to IMCE. This energy reduction comes mainly from using a fast, efficient, in-memory compressor instead of a serial counter in the accumulation phase. In addition, 5.4$\times$ and 9.7$\times$ better energy efficiencies are reported over ReRAM and ASIC accelerators, respectively.\vspace{-0.5em}

\subsection{Performance Estimation}
\begin{figure}[h]
	\centering
	\includegraphics[width=0.47\textwidth]{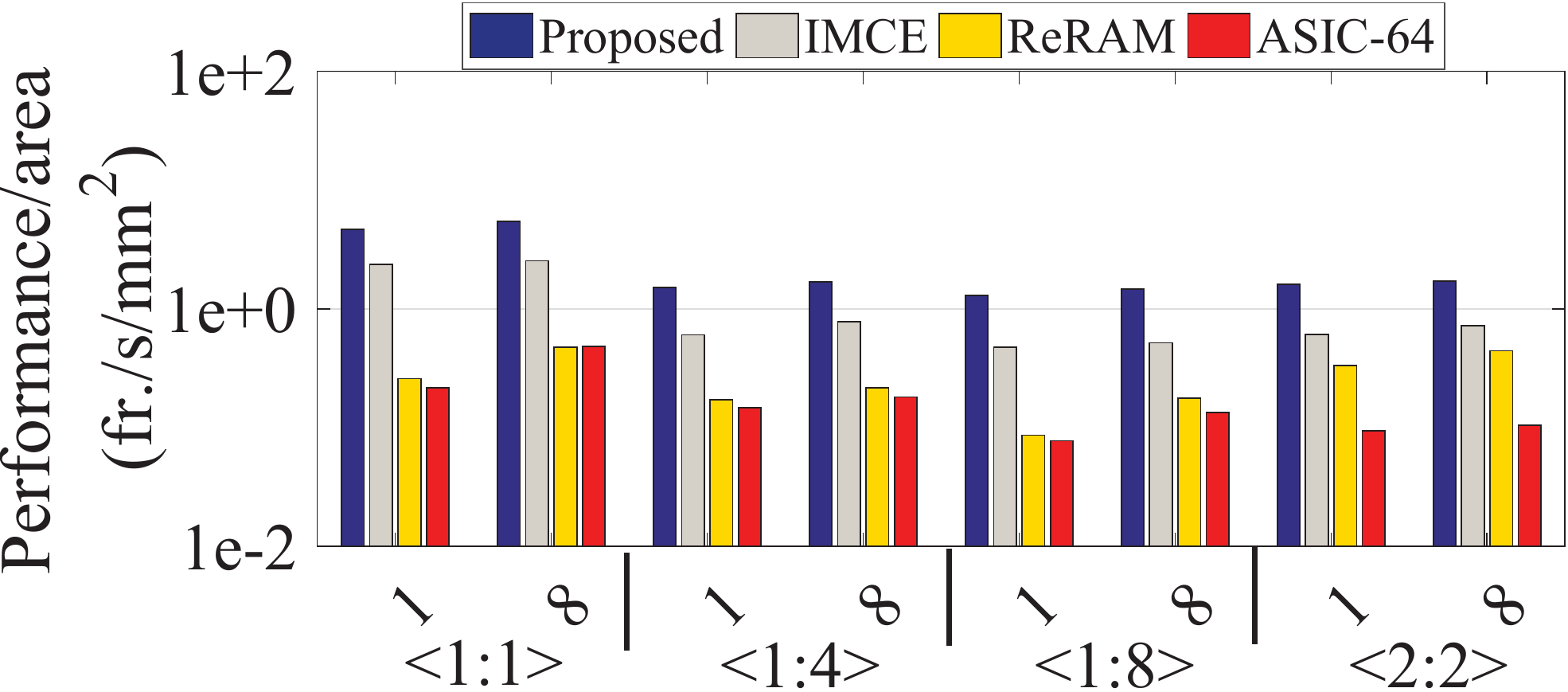}
	\caption{Performance of different accelerators (Y-axis=Log scale).}\vspace{-0.5em}
	\label{perf}	
\end{figure}

Figure \ref{perf} compares the throughput in frames per second normalized to area of the proposed design with other accelerators. We observe that the AND-Accumulation method leads to $\sim$3$\times$ higher performance than AND-bitcount employed in IMCE. In addition, it is 9$\times$ and  13.5$\times$ faster on average than ReRAM and ASIC-64 solutions. This arises from two sources: (1) ultra-fast and parallel in-memory operations of the proposed design compared to multi-cycle ASIC and ReRAM solutions and (2) the existing mismatch between computation and data movement in ASIC design. In addition, the ReRAM design uses matrix splitting approach because of the intrinsically limited bit levels of ReRAM devices so that excessive sub-arrays are occupied. This can further limit parallelism methods  \cite{chi2016prime}.\vspace{-0.7em}

\subsection{Area-Energy trade-off}
In this subsection, we evaluate the energy/area of BCNN resistive processing-in-memory accelerators based on ReRAM \cite{tang2017binary} and SOT-MRAM \cite{angizi2018imce}) for inference of one single image over three well-known data-sets under a 45nm technology node. Table \ref{AreaEnergy} demonstrates that the proposed SOT-MRAM-based accelerator can process BCNN very efficiently compared to others. Its worth pointing out that the energy reported in Table \ref{AreaEnergy} consists of the energy of convolution computation of all layers.  We observe that our design can execute binary-weight AlexNet \cite{rastegari2016xnor} on ImageNet favorably with 471.8$\mu$J/img where $\sim$4.8$\times$ and 3.5$\times$ smaller energy and area are obtained, respectively, compared to the ReRAM-based design. In addition, the proposed accelerator exhibits 1.6$\times$ better energy savings compared to SOT-MRAM IMCE on ImageNet even though it imposes larger overhead to the memory chip.

\begin{table}[h]
\centering
\caption{Energy-area comparison of different NVM-based BCNN accelerators.}
\label{AreaEnergy}
\scalebox{0.90}{
\begin{tabular}{c|c|c|c|c|c|c|}
\cline{2-7}
                                                                                                 & \multicolumn{2}{c|}{ImageNet}                                                                                            & \multicolumn{2}{c|}{SVHN}                                                                                                & \multicolumn{2}{c|}{MNIST}                                                                                              \\ \hline
\multicolumn{1}{|c|}{Designs}                                                                    & \begin{tabular}[c]{@{}c@{}}Energy \\ ($\mu$J/img)\end{tabular} & \begin{tabular}[c]{@{}c@{}}Area\\ ($mm^2$)\end{tabular} & \begin{tabular}[c]{@{}c@{}}Energy\\  ($\mu$J/img)\end{tabular} & \begin{tabular}[c]{@{}c@{}}Area\\ ($mm^2$)\end{tabular} & \begin{tabular}[c]{@{}c@{}}Energy\\ ($\mu$J/img)\end{tabular} & \begin{tabular}[c]{@{}c@{}}Area\\ ($mm^2$)\end{tabular} \\ \hline
\multicolumn{1}{|c|}{\begin{tabular}[c]{@{}c@{}}ReRAM \\ \cite{tang2017binary}\end{tabular}} & 2275.34                                                        & 9.19                                                    & 425.21                                                         & 0.085                                                   & 13.55                                                         & 0.060                                                   \\ \hline
\multicolumn{1}{|c|}{\begin{tabular}[c]{@{}c@{}}IMCE \\ \cite{angizi2018imce}\end{tabular}}                                                             & 785.25                                                         & 2.12                                                    & 135.26                                                          & 0.01                                                   & 0.92                                                          & 0.009                                                  \\\hline

\multicolumn{1}{|c|}{Proposed}                                                             &     471.8                                                     &    2.60                                                &   84.31                                                      &  0.039                                                &     0.68                                                     &  0.012                                                \\
\hline
\end{tabular}}
\vspace{-1.5em}
\end{table}

\section{Conclusion}\vspace{-0.4em}
In this work, a bit-wise CNN in-memory accelerator based on SOT-MRAM computational sub-arrays was proposed. This new architecture could be leveraged to greatly reduce energy consumption dealing with convolutional layers and accelerate low bit-width CNN inference within non-volatile MRAM. Our device-to-architecture co-simulation results show that the proposed accelerator can attain $\sim$5.4$\times$ higher energy-efficiency and 9$\times$ speedup compared to ReRAM-based, and, $\sim$9.7$\times$ higher energy-efficiency and 13.5$\times$ speedup over ASIC accelerators holding almost the same inference accuracy to the baseline CNN on different data-sets. 

We plan to extend our future work to mitigate the write-operations issue for NV elements, which consumes a large amount of power. Choosing a proper thermal barrier, i.e. 30kT, for MTJ devices could provide retention times ranging from minutes to hours and achieve at least 50\% energy reduction compared to nanomagnets with a thermal barrier around 40kT. The other approach to reduce performance overhead caused by NV elements is to leverage one NV-FF instead of two NV-FFs within each FAs. After a specified duration, only Cout will be stored in a NV-FF while sum is saved in a regular FF. If power failure occurs, the stored value is considered as both sum and Cout for the next add operation. In this scenario, PDP improvements can be achieved at the cost of lower accuracy. Generally, in a situation with a high occurrence rate of power failure, the number of completed tasks for a CMOS-only implementation is significantly reduced, which degrades performance of the system \cite{Roohi2018TC}. Hence, utilizing the power failure resilient architecture even without further optimization can avoid high bulk-write energy costs of Flash pages and complexities from checkpointing/restore protocols.\vspace{-0.5em}

\vspace{1em}

\bibliographystyle{IEEEtran}  
\footnotesize
\bibliography{IEEEabrv,./Reference}
\balance

\end{document}